\documentclass[acmtog,screen,nonacm]{acmart} 
\acmSubmissionID{***}

\usepackage{booktabs} 
\usepackage{multirow,makecell}
\usepackage{paralist}
\usepackage{enumitem}
\usepackage[capitalize,nameinlink]{cleveref}
\crefname{section}{Sec.}{Secs.}
\Crefname{section}{Section}{Sections}
\Crefname{table}{Table}{Tables}
\crefname{table}{Tab.}{Tabs.}

\usepackage[ruled]{algorithm2e} 

\SetAlFnt{\small}
\SetAlCapFnt{\small}
\SetAlCapNameFnt{\small}
\SetAlCapHSkip{0pt}

\acmJournal{TOG}

\newcommand{\OURS}{GANeRF}
\definecolor{cyan}{RGB}{0, 255, 255}
\definecolor{magenta}{RGB}{255, 0, 127}

\begin{document}
\title{\OURS: Leveraging Discriminators to Optimize Neural Radiance Fields}

\author{Barbara Roessle}
\affiliation{%
  \institution{Technical University of Munich}
  \country{Germany}}
\author{Norman M{\"u}ller}
\affiliation{%
  \institution{Technical University of Munich}
  \country{Germany}}
\affiliation{%
  \institution{Meta Reality Labs Zurich}
  \country{Switzerland}}
\author{Lorenzo Porzi}
\affiliation{%
  \institution{Meta Reality Labs Zurich}
  \country{Switzerland}}
\author{Samuel Rota Bul{\`o}}
\affiliation{%
  \institution{Meta Reality Labs Zurich}
  \country{Switzerland}}
\author{Peter Kontschieder}
\affiliation{%
  \institution{Meta Reality Labs Zurich}
  \country{Switzerland}}
\author{Matthias Nie{\ss}ner}
\affiliation{%
  \institution{Technical University of Munich}
  \country{Germany}}
\begin{abstract}
Neural Radiance Fields (NeRF) have shown impressive novel view synthesis results; nonetheless, even thorough recordings yield imperfections in reconstructions, for instance due to poorly observed areas or minor lighting changes.
Our goal is to mitigate these imperfections from various sources with a joint solution: we take advantage of the ability of generative adversarial networks (GANs) to produce realistic images and use them to enhance realism in 3D scene reconstruction with NeRFs. 
To this end, we learn the patch distribution of a scene using an adversarial discriminator, which provides feedback to the radiance field reconstruction, thus improving realism in a 3D-consistent fashion. 
Thereby, rendering artifacts are repaired directly in the underlying 3D representation by imposing multi-view path rendering constraints. 
In addition, we condition a generator with multi-resolution NeRF renderings which is adversarially trained to further improve rendering quality. 
We demonstrate that our approach significantly improves rendering quality, e.g., nearly halving LPIPS scores compared to Nerfacto while at the same time improving PSNR by 1.4dB on the advanced indoor scenes of Tanks and Temples.
\end{abstract}

%
%

%
%


\setcopyright{acmlicensed}
\acmJournal{TOG}
\acmYear{2023} \acmVolume{42} \acmNumber{6} \acmArticle{} \acmMonth{12} \acmPrice{15.00}\acmDOI{10.1145/3618402}

\begin{teaserfigure}
  \includegraphics[width=\textwidth,trim={0.cm 9.6cm 0.cm .cm},clip]{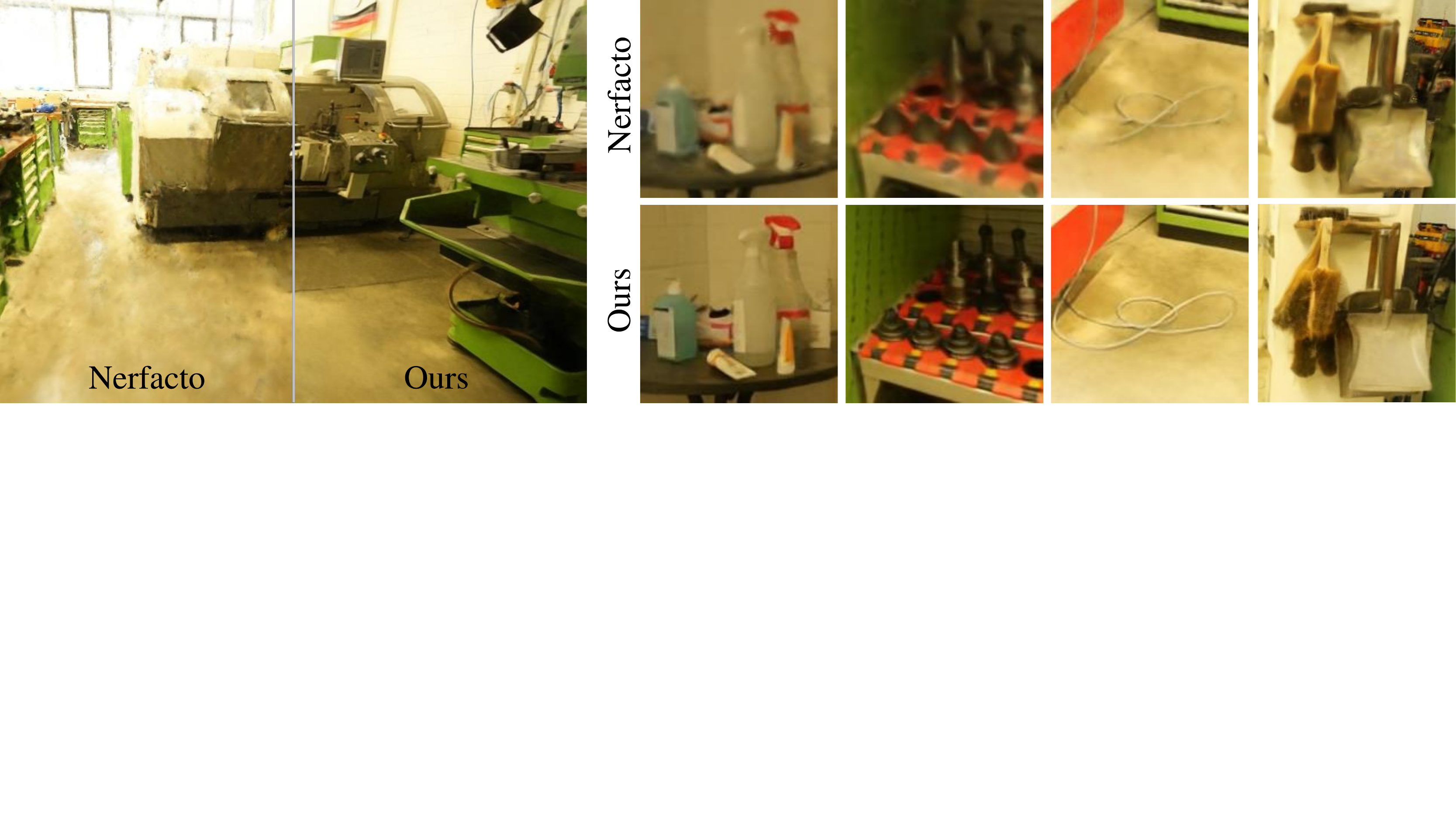}
  \caption{GANeRF proposes an adversarial formulation whose gradients provide feedback for a 3D-consistent radiance field representation. This introduces additional constraints that enable more realistic renderings, and lead to improved novel view synthesis compared to Nerfacto and other baselines.}
  \label{fig:teaser}
\end{teaserfigure}

\maketitle

\section{Introduction}
Neural Radiance Fields (NeRFs)~\cite{mildenhall2020nerf} can achieve remarkable novel view synthesis (NVS) results, powering applications in the domains of virtual/mixed reality, robotics, computational photography, and many others. 
Given a set of posed input images, NeRFs distill complex and viewpoint-dependent scene information, parameterized as 5D input vectors (3D coordinates + 2D viewing direction), into volumetric density and color fields modeled with a neural network.
Volumetric rendering techniques are then applied to generate photorealistic 2D output images for novel camera views from these fields. 
While NeRFs are highly effective at providing compact scene representations that enable photorealistic NVS, their applicability is nonetheless still limited for in-the-wild use cases.

That is, NeRFs are optimized to overfit to the training dataset’s appearance information, which makes them highly dependent on carefully collected input data and properly chosen regularization strategies. In fact, the shape-radiance ambiguity~\cite{kaizhang2020} describes that a set of training images can be perfectly regenerated from a NeRF without respecting the underlying geometry, but by simply exploiting the view-dependent radiance to simulate the actual scene geometry. As a consequence, novel view synthesis quality for non-training camera views drastically degrades and the generated images exhibit well-known cloudy \textit{floater} artifacts. Unfortunately, even with significantly increased capture efforts and when input data comprises dense coverage for a scene, the reconstruction problem can remain ambiguous in areas with changing lighting conditions or in reflective or low-textured regions. Finally, for making NeRFs more widely adopted and applicable, the image capture process needs to be simple and of low effort while yielding high-quality reconstruction results.

In our work, we take advantage of generative adversarial networks (GANs) to improve the NeRF quality in challenging real-world scenarios. 
To this end, we introduce GANeRF -- a novel approach for resolving imperfections directly in the NeRF reconstruction.
Our key idea is to leverage an adversarial loss formulation  in an end-to-end fashion to introduce additional rendering constraints from a per-scene 2D discriminator.
In particular, in regions with limited observations, this enforces the radiance field representation to generate patches that follow the distribution of real-world image patches more closely.
Consequently, \OURS~enables notable mitigation of quality degradation effects in NVS due to imperfect input data, independently of their root causes like limited coverage, image distortion, or illumination changes.

We propose a joint adversarial training during the NeRF optimization, such that a 2D patch discriminator informs the NeRF about the degree of photorealism for rendered patches. 
Through gradient feedback into the 3D scene representation, we reduce typical imperfections in the radiance field reconstruction while inherently encouraging 3D-consistent, photorealistic NeRF renderings.
We show how to further improve the output quality with a subsequent generator that operates on the 2D NeRF renderings at multiple scales, by refining them to provide closer matches to the real distribution of the scene's images. 
We evaluate our method on challenging indoor scenes from the novel ScanNet++~\cite{scannetpp} and the well-known Tanks and Temples~\cite{knapitsch2017tanks} datasets, and show that leveraging our adversarial formulation within NeRFs leads to significant image quality improvements over prior works. 
Across all test scenes, we obtain remarkable improvements over the best-performing baselines~\cite{barron2022mipnerf360,nerfstudio} for perceptual metrics like LPIPS (reductions between 28-48\%), while maintaining consistently better PSNR and SSIM scores.

\smallskip
In summary, we provide the following contributions:
\begin{itemize}
    \item We introduce a novel adversarial formulation that imposes patch-based rendering constraints obtained from a 2D discriminator to optimize a 3D-consistent radiance field representation.
    \item We propose a 2D generator that further refines the rendering output, demonstrating significant improvements over state-of-the-art methods in novel view synthesis on challenging, large-scale scenes.
\end{itemize}
\section{Related Work}
\begin{figure*}[tb]
  \centering
\includegraphics[width=0.89\textwidth,trim={0.3cm 2.6cm 0.cm 1.cm},clip]{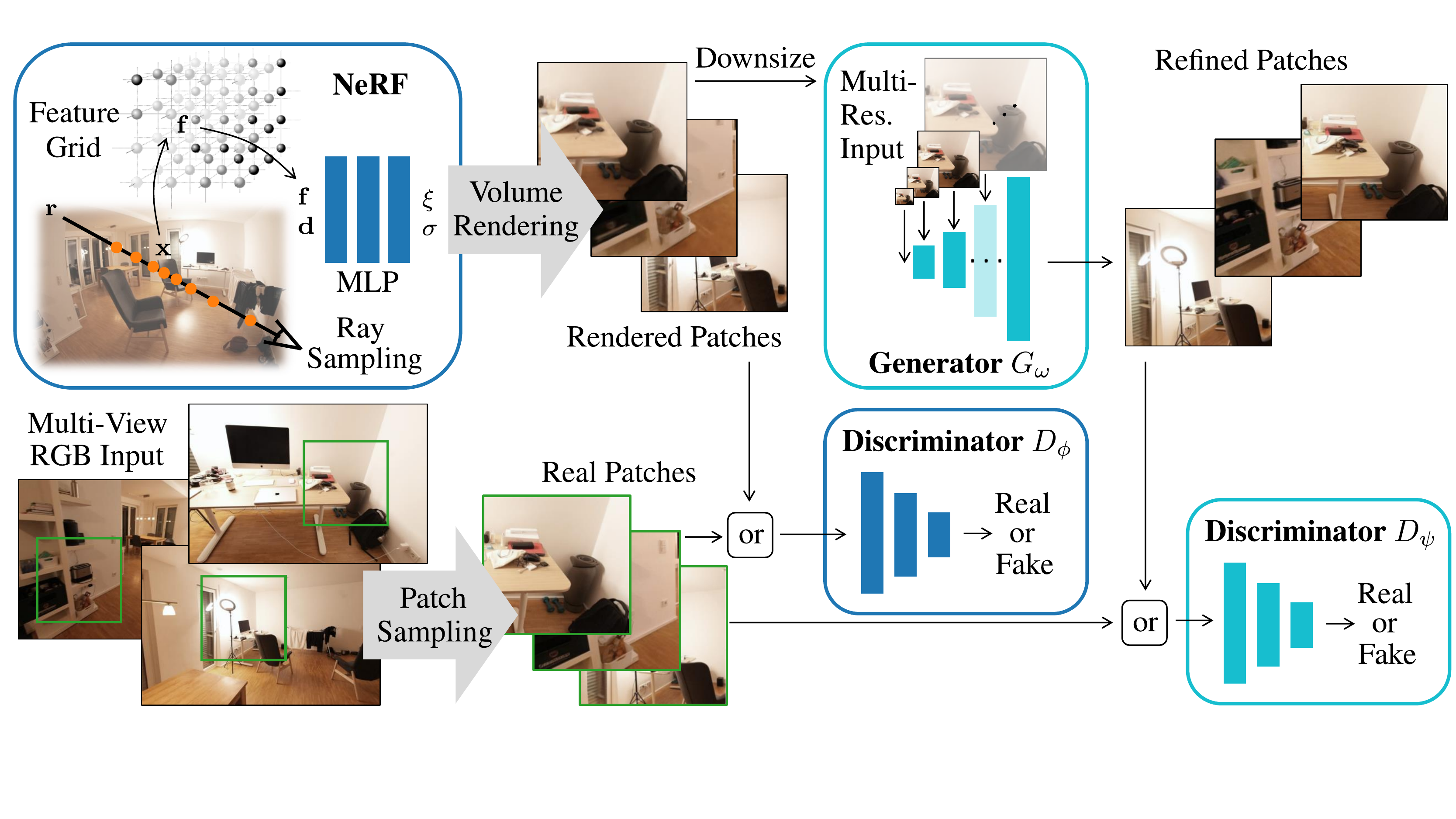}
   \caption{GANeRF method overview: our method takes as input a set of posed images and optimizes for a 3D radiance field representation. Our core idea is to incorporate multi-view patch-based re-rendering constraints in an adversarial formulation that guides the NeRF reconstruction process, and to refine rendered images using a conditional generator network. Particularly in under-constrained regions this significantly improves the resulting rendering quality.}
   \label{fig:pipeline}
\end{figure*}

Neural Radiance Fields (NeRFs) model a 3D scene as a volumetric function, which can be rendered from arbitrary viewpoints to generate highly-realistic images.
While the seminal paper of~\cite{mildenhall2020nerf} encoded this function as a Multi-Layer Perceptron (MLP), more recent works have proposed alternative representations based on spatial data structures such as voxel grids~\cite{yu_and_fridovichkeil2021plenoxels,DirectVoxGO}, plane-based factorizations~\cite{TensoRF}, or multi-scale 3D hash grids~\cite{mueller2022instant}.
Mip-NeRF~\cite{barron2021mipnerf} addressed aliasing issues by introducing an alternative rendering formulation based on conical frustums instead of rays, which was further expanded in~\cite{barron2022mipnerf360} to account for unbounded scenes, and adapted to hash grid-based representations in~\cite{barron2023zip}.
In our work, we follow the Nerfacto model of~\cite{nerfstudio}, which combines the field architecture from Instant-NGP~\cite{mueller2022instant} with the multi-stage proposals of MipNeRF-360~\cite{barron2022mipnerf360} to achieve a good trade-off between speed and quality.

\subsection{NeRFs with Priors}
Although the approaches discussed in the previous section achieve impressive visual fidelity, they generally struggle to represent under-constrained scene regions.
Most NeRFs incorporate one or more heuristics to combat common artifacts such as ``floaters'', e.g., by adding losses that promote peaked density~\cite{barron2021mipnerf,barron2022mipnerf360,hedman2021baking}, injecting noise into the model during training~\cite{mildenhall2020nerf}, or promoting surface smoothness~\cite{wang2021neus, zhang2021nerfactor,oechsle2021unisurf}.
When data is insufficient or ambiguous, however, more effective priors are necessary to regularize the model.

In the few-shot setting, geometric priors have been shown to be particularly effective: pre-trained models can be used to supervise the NeRF with predicted depth~\cite{kangle2021dsnerf, roessle2022depthpriorsnerf} or surface normals~\cite{yu2022monosdf}.
RegNeRF~\cite{Niemeyer2021Regnerf} utilizes a combination of geometric (surface smoothness, sampling space annealing) and appearance (normalizing flow) regularizers to enable training a NeRF on as few as 3 images.
Similarly, SinNeRF~\cite{xu2022sinnerf} proposes a semi-supervised learning approach using geometric and semantic pseudo labels to guide a NeRF reconstruction from a single view.
In our work, we focus on scenes with denser coverage, where geometric priors have been observed to provide only marginal improvements~\cite{Niemeyer2021Regnerf}.

Other works focus on appearance-based priors, e.g., DiffusioNeRF~\cite{wynn-2023-diffusionerf} uses a 2D denoising diffusion model trained on RGBD images to construct an unsupervised loss term, which encourages the NeRF to render plausible images from unobserved viewpoints.
Finally, some approaches learn scene-based priors, by casting NeRF reconstruction as a generalization problem: MVSNeRF~\cite{mvsnerf} constructs a cost volume that is encoded to a neural volume, allowing consistent renderings from only a few images.
PixelNeRF~\cite{yu2020pixelnerf} and GenVS~\cite{chan2023genvs} train an image encoder to lift few input views to 3D-aware neural representations that can be rendered from novel views.
However, these methods are still limited by the availability of training data: approaches based on scene-based priors in particular tend to produce results that lack detail, or impose strong assumptions on scene content and rendering trajectory according to the types of scenes they were trained on.
In contrast, our per-scene adversarial optimization approach avoids the need for any external data.

\subsection{GANs for Image Refinement}
Generative Adversarial Networks (GANs)~\cite{Goodfellow2014gan,Mescheder2018ICML,Karras2019stylegan2} are trained to produce images from a given data distribution by optimizing an adversarial loss.
Initially proposed for unconditional image generation, GANs have also been applied to image-to-image translation and refinement tasks~\cite{pix2pix2017,wang2018pix2pixHD,park2019semantic}, such as colorization~\cite{anwar2020image}, super-resolution~\cite{ledig2017photo}, and in-painting~\cite{elharrouss2020image}.
In a similar setting, we optimize a conditional adversarial formulation based on StyleGAN2~\cite{Karras2019stylegan2} to refine the images produced by a NeRF to more closely match the data distribution of a given scene. 
Prior work has combined GANs with NeRFs for different settings, such as unkown camera poses~\cite{meng2021gnerf}, generative NeRFs of objects and compositions~\cite{Schwarz2020NEURIPS,Niemeyer2020GIRAFFE}, 3D-aware generators~\cite{epigraf,kwak2022injecting} and scene generation~\cite{son2023singraf}. 4K-NeRF~\cite{wang20224k} combines a low resolution NeRF with a 3D-aware decoder for super-resolution. In our work, we found that direct backpropagation from a discriminator to a full resolution NeRF is essential for view-consistent novel view synthesis.  
A related idea is explored concurrently to our work in NeRFLiX~\cite{zhou2023nerflix}, which trains a network in a non-adversarial setting to invert NeRF artifacts.
In contrast, the approach in~\cite{zhou2023nerflix} requires training on multiple scenes and relies on a hand-crafted model to simulate NeRF noise. 

\section{Method}
Given a set of posed input images capturing a static 3D scene, we focus on the problem of synthesizing novel views of that scene.
We build on top of recent Neural Radiance Fields (NeRF)~\cite{mildenhall2020nerf}, specifically, the Nerfacto model of~\cite{nerfstudio}. However, our ideas can also be applied to a different NeRF architecture, as shown in the experiments (\cref{sssec:nerf_backbone}). 
To improve realism in novel views, our method (Fig.~\ref{fig:pipeline}) leverages a 2D adversarial loss that directly updates the 3D scene representation.
To this end, a discriminator learns the distribution of image patches in the training data.
Through adversarial training, the 3D NeRF representation (\cref{ssec:nerf_preliminary}) is updated towards rendering patches that match this distribution (\cref{ssec:nerf_opt}).
On top of that, a 2D generator considers NeRF renderings at multiple resolutions, refining them based on feedback from a second discriminator (\cref{ssec:generator}).
\begin{figure*}[tb]
  \centering
\includegraphics[width=\linewidth,trim={1.6cm 12.1cm 1.6cm 0.1cm},clip]{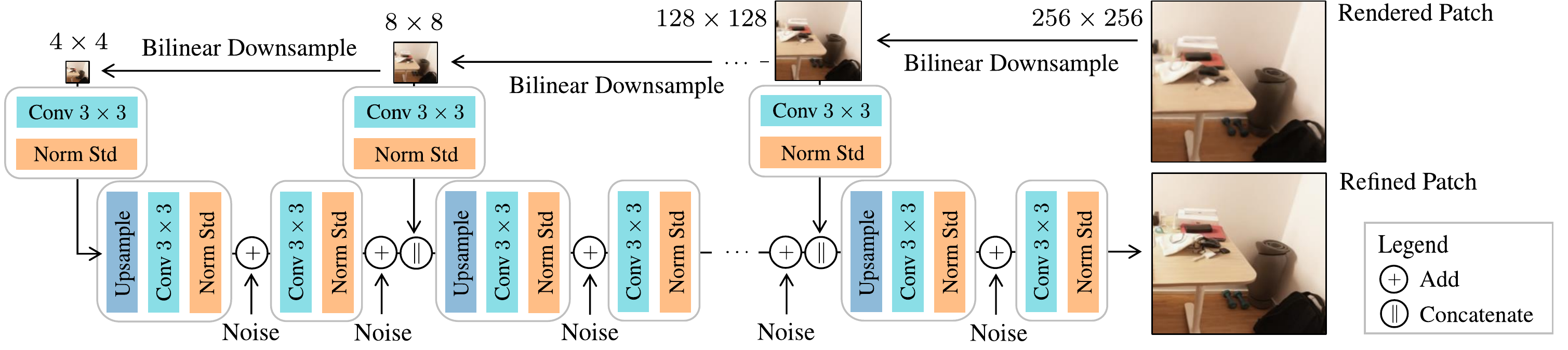}
   \caption{Our conditional generator architecture consists of a feature extraction pyramid of six blocks operating at multiple resolutions: Starting from a $4\times4$ down-sampled patch, convolutional blocks extract level-specific features that are upsampled and added together with generative noise to the next feature extraction level. The final output resolution matches the input resolution of $256\times256$.}
   \label{fig:sg2}
\end{figure*}
\subsection{NeRF Preliminaries}
\label{ssec:nerf_preliminary}
NeRF models represent a scene by providing a density $\sigma_\theta(\mathbf x)$ and an RGB color $\xi_\theta(\mathbf x,\mathbf d)$ for each point $\mathbf x\in\mathbb R^3$ in 3D space, the color depending additionally on the viewing direction $\mathbf d\in\mathbb R^3$ to account for view-dependent effects. This radiance field representation $(\sigma_\theta, \xi_\theta)$, which depends on learnable parameters $\theta$, allows rendering a pixel of an image by casting a ray from the camera origin $\mathbf{o}\in\mathbb R^3$ through the pixel in the direction $\mathbf{d}$ and by computing the expected observed color along the ray with a distribution depending on the density field~\cite{mildenhall2020nerf}. 

Formally, the rendering function for a radiance field $(\sigma_\theta,\xi_\theta)$ and ray $\mathbf r$ is given by
\begin{equation}
\mathcal R_\theta(\mathbf{r})\coloneqq\int_0^\infty \sigma_\theta(\mathbf r_t)\exp\left[-\int_0^t\sigma_\theta(\mathbf r_s)ds\right]\xi_\theta(\mathbf r_t, \mathbf{d})dt\,.
\end{equation}
Here, $\mathbf r_t\coloneqq \mathbf{o} + t\mathbf{d}$ is the 3D point along the ray at time $t$.
$\mathcal R_\theta(\mathbf{r})$ can be approximated with the quadrature rule given samples of density and color along the ray~\cite{mildenhall2020nerf}.

The NeRF parameters $\theta$ are typically optimized to minimize the expected, per-ray mean squared error, which penalizes differences between the rendered color and the ground truth color. Specifically, given ray/color pairs $(\mathbf r,\mathbf{c})$ distributed as $p(\mathbf{r}, \mathbf{c})$, we minimize
\begin{equation}
\mathcal{L}^\mathtt N_{\mathrm{rgb}}(\theta) \coloneqq \mathbb E_p\left\Vert\mathcal R_\theta(\mathbf{r})- \mathbf{c} \right\Vert_2^2\,.
\end{equation}
The superscript $\mathtt N$ distinguishes losses that are used to train NeRF from the ones used to train the generator, which are defined later.

\subsection{NeRF Optimization with Discriminator}
\label{ssec:nerf_opt}
It is often the case that 
scenes are rich in repeated elements and surfaces often look similar even from different perspectives. Accordingly, patches from one view form a good prior for patches in other views. To encode this prior in our training scheme, we complement $\mathcal L_\mathrm{rgb}^\mathtt N$ with an adversarial objective $\mathcal{L}^\mathtt N_\mathrm{adv}(\theta, \phi)$ that leverages an auxiliary neural network $D_\phi$ (\emph{a.k.a.} discriminator) parametrized by $\phi$. The objective is optimized such that the discriminator is pushed towards classifying whether a given image patch is real or produced by the NeRF parametrized by $\theta$ (\emph{a.k.a.} fake), while the NeRF is encouraged to fool the discriminator, thus rendering realistic patches. Patches $P$ are assumed to be distributed according to $q(P)$ and we denote by $\mathbf{r}_P$ the set of rays and by $\mathbf{c}_P$ the pixel colors corresponding to patch $P$. The adversarial objective should be minimized with respect to the NeRF parameters $\theta$ as a proper loss, but maximized with respect to the discriminator parameters $\phi$. Following~\cite{Mescheder2018ICML}, we adopt an $R_1$ gradient penalty term on the discriminator balanced by a nonnegative scalar $\lambda^\mathtt N_{\mathrm{gp}}$ yielding the following form of the adversarial objective:
\begin{multline}
\mathcal{L}^\mathtt N_{\mathrm{adv}}(\theta, \phi)\coloneqq \mathbb{E}_{q}\left[ f(D_\phi(\mathcal R_\theta(\mathbf r_P)))+ f(-D_\phi(\mathbf{c}_P))\right.\\\left.-\lambda^\mathtt N_{\mathrm{gp}}\Vert\nabla D_\phi(\mathbf{c}_P)\Vert_2^2\right]\,.
\end{multline}
Here, by setting $f(x) \coloneqq -\log(1+\exp(-x))$, we obtain a regularized version of the loss originally proposed in the seminal GAN paper~\cite{Goodfellow2014gan}.

We further employ a perceptual loss based on VGG~\cite{Johnson2016perceptual}, which has shown success in conjunction with 2D conditional GANs~\cite{PAN}: 
\begin{equation}
    \mathcal{L}^\mathtt N_{\mathrm{perc}}(\theta) \coloneqq \mathbb E_q\left[\Vert \Phi_\mathtt{VGG}(\mathcal R_\theta(\mathbf r_P))-\Phi_\mathtt{VGG}(\mathbf{c}_P)\Vert_2^2\right]\,,
\end{equation}
where $\Phi_\mathtt{VGG}$ is the vectorized concatenation of the first 5 feature layers before the max pooling operation of a VGG19 network~\cite{vgg}, each layer being normalized by the square root of the number of entries.
$\mathcal{L}^\mathtt N_{\mathrm{perc}}(\theta)$ encourages similarity of real and fake patch features at different granularities. 

The final 
loss that we minimize to train the NeRF is given by
\begin{equation}
    \mathcal{L}_\mathrm N(\theta) \coloneqq \mathcal{L}^\mathtt N_{\mathrm{rgb}}(\theta)+ \lambda^\mathtt N_{\mathrm{perc}}\mathcal{L}^\mathtt N_{\mathrm{perc}}(\theta) + \lambda^\mathtt N_{\mathrm{adv}}\max_{\phi\in\Phi}\mathcal{L}^\mathtt N_{\mathrm{adv}}(\theta,\phi)\,, 
\end{equation}
where $\Phi$ denotes the set of possible discriminators and $\lambda^\mathtt N_*$ are nonnegative balancing factors for the different losses. This objective is optimized with stochastic gradients computed from batches that combine rays and patches sampled uniformly across all training images (i.e., from the $p$ and $q$ distributions, respectively). Moreover, parameter updates are alternated between the NeRF and the discriminator to cope with the saddle-point problem, as usually done in GAN training schemes. 

\subsection{Conditional Generator}
\label{ssec:generator}
Up to this point, artifacts have been repaired directly in 3D through the discriminator guidance introduced in \cref{ssec:nerf_opt}. However, the rendering quality can be further improved in 2D by postprocessing synthesized images. To this end, we design a conditional generator $G_\omega$ parameterized by $\omega$, which takes as input a NeRF rendering and an auxiliary random vector $\mathbf{z}$ and produces a cleaned image, thus serving as a stochastic denoiser. The architecture is inspired by the StyleGAN2 generator~\cite{Karras2019stylegan2}, but we adapt it to a conditional generator and omit the mapping network (\cref{fig:sg2}). We keep the additive noise depending on $\mathbf{z}$, which is close in spirit to the way pix2pix~\cite{pix2pix2017} uses dropout as source of randomness in their conditional GAN setting. 
As shown in \cref{fig:sg2}, the conditioning input patch is 6 times bilinearly downsampled by a factor of 2, thus enabling a multi-scale analysis. The downscaled patches are encoded with a convolutional layer into 32 feature channels, which are concatenated to the output of the generator layer of corresponding scale. 
As in StyleGAN2, we use leaky ReLU activations, input/output skip connections and normalization layers. Additive noise is injected after the normalization layers, and noise is additionally scaled by a learnable per-layer factor.

Akin to the NeRF optimization, the generator is trained in a regularized adversarial setting. This involves a second discriminator $D_\psi$ parameterized by $\psi$, which differs from the previously introduced $D_\phi$. Indeed, the NeRF and the conditional generator introduce different types of errors in the output image that can be addressed more effectively by independent discriminators. The adversarial objective $\mathcal{L}^\mathtt G_{\mathrm{adv}}$ used to train the generator takes the following form:
\begin{multline}
\mathcal{L}^\mathtt G_{\mathrm{adv}}(\omega, \psi|\theta)\coloneqq \mathbb{E}_{q,n}\left[ f(D_\psi(G_\omega(\mathcal R_{\bot\theta}(\mathbf r_P),z)))\right.\\\left.+ f(-D_\psi(\mathbf{c}_P))-\lambda^\mathtt G_{\mathrm{gp}}\Vert\nabla D_\psi(\mathbf{c}_P)\Vert_2^2\right]\,,
\end{multline}
where the expectation is additionally taken with respect to $\mathbf z$ distributed as a standard normal distribution $n(\mathbf z)$. The adversarial objective should be minimized with respect to the generator's parameters $\omega$, but maximized with respect to the discriminator parameter $\psi$. Moreover, we stop the gradient through the NeRF parameters $\theta$, which is denoted by $\bot\theta$.

The adversarial objective is complemented with the same perceptual loss used to train the NeRF, namely
\begin{equation}
    \mathcal{L}^\mathtt G_{\mathrm{perc}}(\omega|\theta) \coloneqq \mathbb E_{q,n}\left[\Vert \Phi_\mathtt{VGG}(G_\omega(\mathcal R_{\bot\theta}(\mathbf r_P),\mathbf{z}))-\Phi_\mathtt{VGG}(\mathbf{c}_P)\Vert_2^2\right]
\end{equation}
and an $L_1$ color loss, as in the conditional GAN pix2pix~\cite{pix2pix2017}, operating on patches, defined as:
\begin{equation}
    \mathcal{L}^\mathtt G_{\mathrm{rgb}}(\omega|\theta) \coloneqq\mathbb E_{q,n}\left[\left\Vert G_\omega(\mathcal R_{\bot\theta}(\mathbf{r}_P), \mathbf{z})- \mathbf{c}_P\right\Vert_1\right]\,.
\end{equation}
The overall generator loss is defined as 
\begin{multline}
        \mathcal{L}_{\mathrm{G}}(\omega|\theta^\star) \coloneqq \lambda^\mathtt G_{\mathrm{perc}}\mathcal{L}_{\mathrm{perc}}^\mathtt G(\omega|\theta^\star) + \lambda^\mathtt G_{\mathrm{rgb}}\mathcal{L}^\mathtt G_{\mathrm{rgb}}(\omega|\theta^\star)\\+\max_{\psi\in\Psi}\mathcal{L}^\mathtt G_{\mathrm{adv}}(\omega,\psi|\theta^\star),     
\end{multline}
where $\Psi$ denotes the set of possible discriminators, $\lambda^\mathtt G{*}$ are nonnegative factors balancing the loss components, and $\theta^\star$ is the parametrization of the NeRF obtained by optimizing $\mathcal L_\mathrm N$.
We cope with the saddle-point problem by alternating updates of the generator and the discriminator akin to what we have described for $\mathcal L_\mathrm N$.

\section{Experiments}
\subsection{Datasets and Metrics}
\subsubsection{ScanNet++}
From the ScanNet++~\cite{scannetpp} dataset, we evaluate on five indoor scenes consisting of office, lab and apartment environments.
Each scene contains an average of $\sim 700$ images, $\sim 20$ of which are defined as a test set.
In contrast to other datasets commonly used in NeRF works (e.g., LLFF~\cite{mildenhall2020nerf} or the 360$^{\circ}$ scenes from~\cite{barron2022mipnerf360}), the test views in ScanNet++ are selected to be spatially out-of-distribution compared to the training views, in order 
to explicitly evaluate a model's generalization capabilities. We use the DSLR images with the provided camera poses from structure from motion~\cite{schoenberger2016sfm}, and undistort and resize them to $768 \times 1152$. 

\subsubsection{Tanks and Temples}
As a benchmark for larger-scale reconstruction, we consider four scenes from the advanced set of scenes of the Tanks and Temples \cite{knapitsch2017tanks} dataset: Auditorium, Ballroom, Courtroom, and Museum.
These scenes depict large indoor environments, with detailed geometries and complex illumination.
For each scene, we randomly select 10\% of the available images as a test set, resulting in a split of $\sim 270$ training and $\sim 30$ test views on average. We use the original resolution of $1080 \times 1920$.

\subsubsection{Evaluation Metrics}
We evaluate our results in terms of three main visual quality metrics: PSNR, SSIM~\cite{wang2004image}, and LPIPS~\cite{zhang2018unreasonable}.
Following previous works in the image generation literature, we additionally report KID~\cite{binkowski2018demystifying} scores, which measure how closely our model's outputs match the visual distribution of the scene.

\subsection{Training and Implementation Details}
\label{ssec:train_details}
\subsubsection{NeRF}
In each training iteration, we render 4096 random rays and a $256 \times 256$ patch from a single random input image.
The perceptual loss processes the patch as a whole, whereas the adversarial loss subdivides it into 16 smaller patches of size $64\times 64$ without overlap.
$D_\phi$ is a StyleGAN2 discriminator~\cite{Karras2019stylegan2}, which is adapted to process $64 \times 64$ patches with the number of convolutional channels reduced by half. 
For each generated ``fake'' patch the discriminator is also given a real patch.
The radiance field and discriminator $D_\phi$ are trained for 400k iterations, using the Adam optimizer~\cite{Kingma2015AdamAM} and RMSprop~\cite{rmsprop} with learning rates $1\times10^{-2}$ and $1\times10^{-3}$, respectively. 
The loss weights are set to $\lambda^\mathtt N_{\mathrm{adv}}=0.0003$, $\lambda^\mathtt N_{\mathrm{perc}}=0.0003$ and $\lambda^\mathtt N_{\mathrm{gp}}=0.1$. 

\subsubsection{Conditional Generator}
The initial training patch size before downsampling is $256 \times 256$. 
At inference time, however, the generator is able to run fully convolutionally on high-resolution images to refine the NeRF renderings. 
The generator is trained using a batch size of 8. While the perceptual loss operates on the full training patches, the adversarial loss subdivides each patch into 4 smaller patches of size $128 \times 128$ without overlap, thus $D_\psi$ is trained with batch size 32. The discriminator is inspired by StyleGAN2~\cite{Karras2019stylegan2}, and adapted to process $128 \times 128$ patches. 
Both generator and discriminator $D_\psi$ are trained for 3000 epochs using the Adam optimizer~\cite{Kingma2015AdamAM} with learning rate $2\times10^{-3}$. 
The loss weights are set to $\lambda^\mathtt G_{\mathrm{perc}}=1.0$, $\lambda^\mathtt G_{\mathrm{rgb}}=3.0$ and $\lambda^\mathtt G_{\mathrm{gp}}=5.0$. 

\subsection{Baseline Comparisons}
\begin{figure*}[tbp]
 \setlength\tabcolsep{0pt}
  \centering
  \begin{tabular}{>{\centering\arraybackslash}p{0.1666\textwidth}>{\centering\arraybackslash}p{0.1666\textwidth}>{\centering\arraybackslash}p{0.1666\textwidth}>{\centering\arraybackslash}p{0.1666\textwidth}>{\centering\arraybackslash}p{0.3333\textwidth}}
\multicolumn{5}{c}{\includegraphics[width=\textwidth,trim={-0cm 0cm 0.cm 0cm},clip]{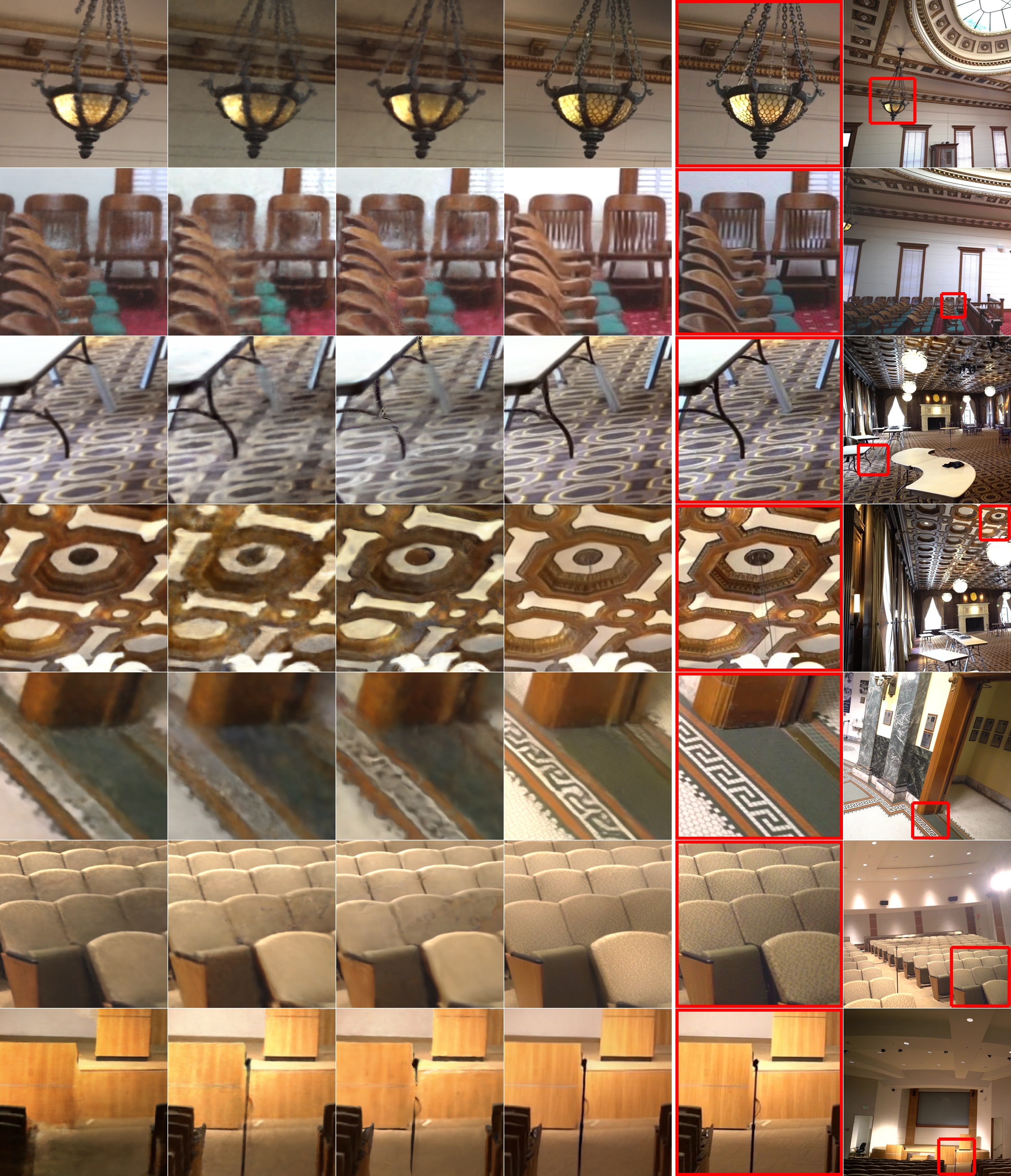}} \\
Mip-NeRF 360 & Instant NGP & Nerfacto & Ours & Ground truth \\
  \end{tabular}
  \vspace{-0.2cm}
   \caption{Comparison on Tanks and Temples. Our method recovers more detail than the baselines, such as thin structures or patterns on the floor. 
   }
   \label{fig:tnt_baseline_comp}
\end{figure*}

\begin{figure*}[tbp]
 \setlength\tabcolsep{0pt}
  \centering
  \begin{tabular}{>{\centering\arraybackslash}p{0.1666\textwidth}>{\centering\arraybackslash}p{0.1666\textwidth}>{\centering\arraybackslash}p{0.1666\textwidth}>{\centering\arraybackslash}p{0.1666\textwidth}>{\centering\arraybackslash}p{0.3333\textwidth}}
\multicolumn{5}{c}{\includegraphics[width=\textwidth,trim={-0cm 0cm 0.cm 0cm},clip]{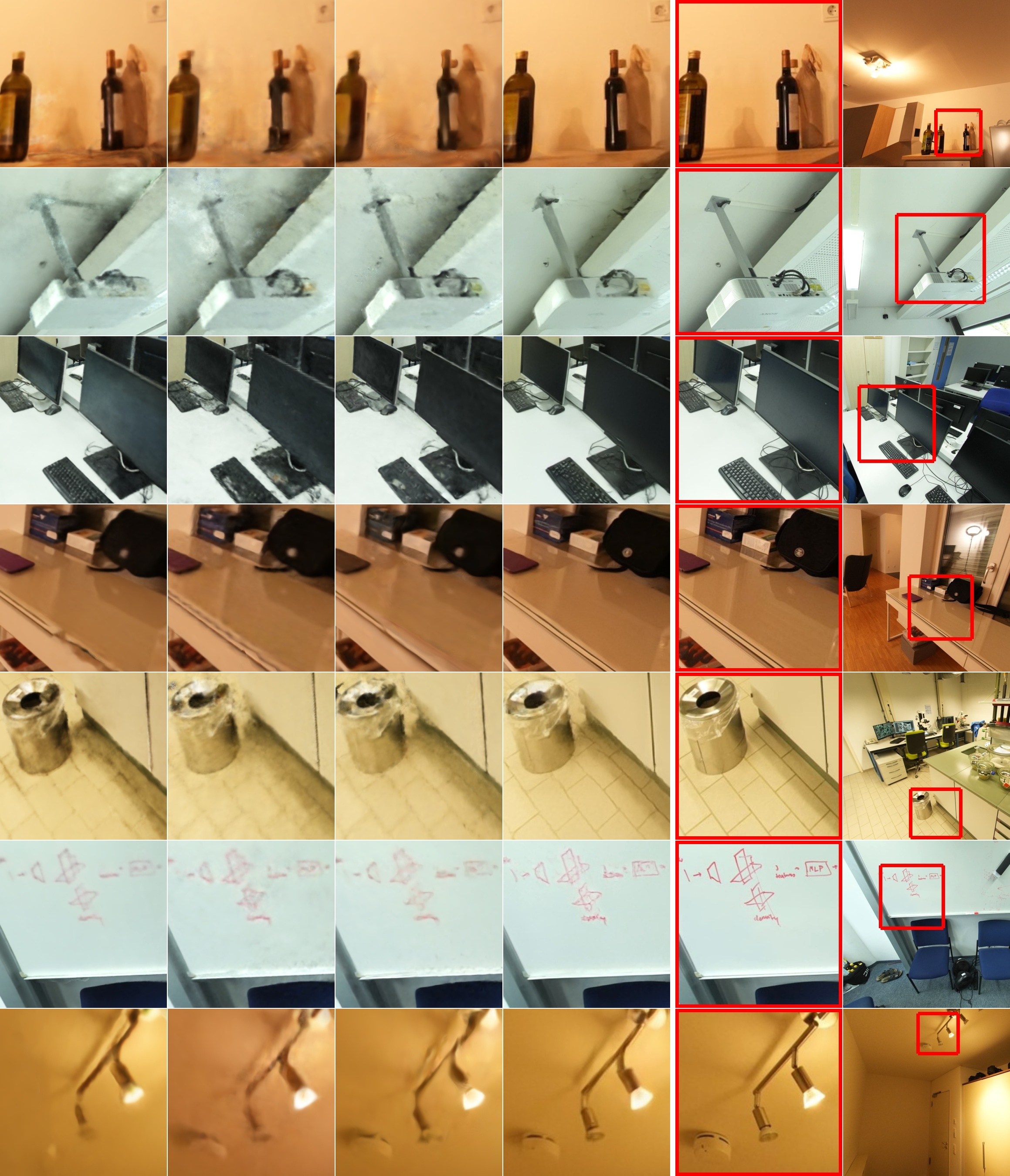}} \\
Mip-NeRF 360 & Instant NGP & Nerfacto & Ours & Ground truth \\
  \end{tabular}
  \vspace{-0.2cm}
   \caption{Comparison on Scannet++. Our method produces less foggy artifacts, which leads to sharper renderings compared to the baselines.
   }
   \label{fig:scannet_baseline_comp}
\end{figure*}

\begin{figure*}[tb]
 \setlength\tabcolsep{0pt}
  \centering
  \begin{tabular}{>{\centering\arraybackslash}p{0.1666\textwidth}>{\centering\arraybackslash}p{0.1666\textwidth}>{\centering\arraybackslash}p{0.1666\textwidth}>{\centering\arraybackslash}p{0.1666\textwidth}>{\centering\arraybackslash}p{0.3333\textwidth}}
\multicolumn{5}{c}{\includegraphics[width=\textwidth,trim={-0cm 0cm 0.cm 0cm},clip]{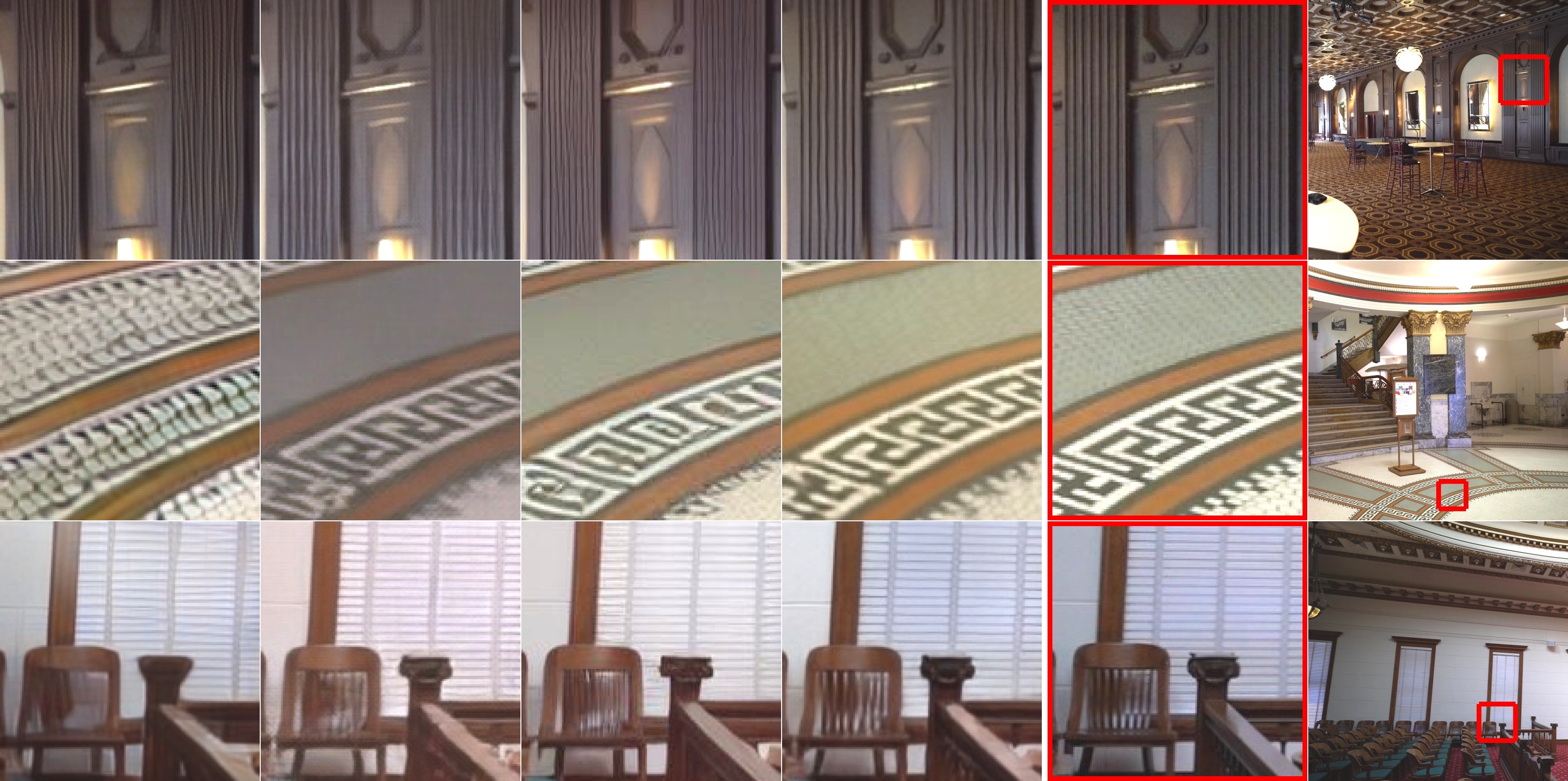}} \\
4K-NeRF & Nerfacto + pix2pix & Nerfacto + ControlNet & Ours & Ground truth \\
  \end{tabular}
  \vspace{-0.2cm}
   \caption{Comparison on Tanks and Temples. Using backpropagation to the NeRF representation and effective generator conditioning, our novel views closely match the ground truth patterns compared to other approaches based on generative models.}
   \label{fig:tnt_baseline_comp_gen}
\end{figure*}

We compare our approach to several baselines from recent literature.
We pick Mip-NeRF 360~\cite{barron2022mipnerf360} and Instant NGP~\cite{mueller2022instant} as representatives of the current state of the art 
in NeRF architectures optimized for visual quality and speed, respectively.
Furthermore, we evaluate four variations of the Nerfacto~\cite{nerfstudio} model that our proposed approach is based upon: 
\begin{compactenum}[i]
    \item Nerfacto: the baseline model with no modifications.
    \item Nerfacto + extra capacity: this higher-capacity variation doubles the number of hidden dimensions of the MLPs, doubles the grid resolution, and increases the hash table size by 4 compared to the default Nerfacto. This results in  $\sim 44M$ trainable parameters, i.e., $\sim 25\%$ more than our model (NeRF + Generator) and $\sim 3.4\times$ the number of parameters of the default Nerfacto ($\sim 12.9M$). 
    \item Nerfacto + pix2pix: Nerfacto results are refined by a pix2pix~\cite{pix2pix2017} generator. From the official pix2pix repository, we found that the generator using 9 ResNet~\cite{resnet} blocks, trained with Wasserstein objective and gradient penalty~\cite{wgangp} performs best. We equally train this model with the VGG perceptual loss, which improves its performance. 
    \item Nerfacto + ControlNet: Nerfacto results are refined by Stable Diffusion~\cite{rombach2022highresolution} with ControlNet~\cite{zhang2023adding} conditioning on Nerfacto renderings and LoRA~\cite{hu2022lora} fine-tuning using the implementation of \cite{wu2023controllora}.
\end{compactenum}
We further compare to the recent work 4K-NeRF~\cite{wang20224k}. For a fair comparison, all baselines are trained until convergence. 
As shown in \cref{tab:results_scannet} and \cref{tab:results_tnt}, our approach leads to noticeable improvements when compared to baselines. 
In particular, the two perceptual metrics (LPIPS, KID) demonstrate the largest relative improvements, suggesting that our approach is able to fix many of the small visual artifacts that are often poorly measured by color similarity metrics (PSNR, SSIM). 
This is also confirmed by the qualitative evaluation shown in \cref{fig:scannet_baseline_comp,fig:tnt_baseline_comp,fig:tnt_baseline_comp_gen}. 
Nerfacto + ControlNet achieves high perceptual quality (LPIPS, KID), however, the results are less view-consistent (see \cref{sssec:consistency}), which also reflects in lower PSNR. 

\begin{table}[tb]
  \centering
  \caption{Quantitative results on five ScanNet++ scenes. 
  Our method outperforms the baselines by a large margin in the perceptual metrics, like LPIPS and KID, while maintaining consistently better PSNR and SSIM scores.}
  \label{tab:results_scannet}
\resizebox{\linewidth}{!}{
  \begin{tabular}{@{}lcccc@{}}
    \toprule
    Method & PSNR$\uparrow$ & SSIM$\uparrow$ & LPIPS$\downarrow$ & KID $\downarrow$ \\
    \midrule
    Mip-NeRF 360 \cite{barron2022mipnerf360} & 24.9 & 0.862 & 0.225 & 0.0241 \\
    Instant NGP \cite{mueller2022instant} & 25.3 & 0.844 & 0.269 & 0.0511 \\
    4K-NeRF \cite{wang20224k} & 22.7 & 0.807 & 0.254 & 0.0350 \\
    Nerfacto \cite{nerfstudio} & 25.6 & 0.848 & 0.245 & 0.0398 \\
    Nerfacto + extra capacity & 25.9 & 0.854 & 0.228 & 0.0314 \\
    Nerfacto + pix2pix \cite{pix2pix2017} & 24.9 & 0.848 & 0.193 & 0.0162 \\
    Nerfacto + ControlNet \begin{minipage}{0.21\linewidth}\small \raggedright \linespread{0.}\selectfont \cite{zhang2023adding}\end{minipage} & 23.1 & 0.827 & 0.174 & \textbf{0.0097} \\
    \hline 
    Ours w/o discriminator & 25.8 & 0.857 & 0.177 & 0.0143 \\
    Ours w/o generator & 25.9 & 0.860 & 0.198 & 0.0169 \\
    Ours & \textbf{26.1} & \textbf{0.864} & \textbf{0.161} & 0.0113 \\
    \bottomrule
  \end{tabular}
  }
\end{table}

\begin{table}[tb]
  \centering
  \caption{Quantitative results on four Tanks and Temples scenes. Our method achieves particularly strong improvements on the perceptual metrics, thus improving visual details and sharpness of novel view renderings.} 
  \label{tab:results_tnt}
\resizebox{\linewidth}{!}{
  \begin{tabular}{@{}lcccc@{}}
    \toprule
    Method & PSNR$\uparrow$ & SSIM$\uparrow$ & LPIPS$\downarrow$ & KID $\downarrow$ \\
    \midrule
    Mip-NeRF 360 \cite{barron2022mipnerf360} & 18.5 & 0.709 & 0.327 & 0.0277 \\
    Instant NGP \cite{mueller2022instant} & 19.3 & 0.700 & 0.369 & 0.0466 \\
    4K-NeRF \cite{wang20224k} & 19.4 & 0.656 & 0.356 & 0.0353 \\
    Nerfacto \cite{nerfstudio} & 19.5 & 0.716 & 0.329 & 0.0432 \\
    Nerfacto + extra capacity & 19.6 & 0.733 & 0.291 & 0.0314 \\
    Nerfacto + pix2pix \cite{pix2pix2017} & 20.6 & 0.739 & 0.242 & 0.0115 \\
    Nerfacto + ControlNet \begin{minipage}{0.21\linewidth}\small \raggedright \linespread{0.}\selectfont \cite{zhang2023adding}\end{minipage} & 19.6 & 0.706 & 0.213 & 0.0085 \\
    \hline 
    Ours w/o discriminator & 20.6 & 0.745 & 0.192 & 0.0102 \\
    Ours w/o generator & 19.9 & 0.739 & 0.251 & 0.0130 \\
    Ours & \textbf{20.9} & \textbf{0.776} & \textbf{0.169} & \textbf{0.0065} \\
    \bottomrule
  \end{tabular}
  }
\end{table}

\begin{figure*}[tbp]
 \setlength\tabcolsep{0pt}
  \centering
  \begin{tabular}{>{\centering\arraybackslash}p{0.04\textwidth}>{\centering\arraybackslash}p{0.16\textwidth}>{\centering\arraybackslash}p{0.16\textwidth}>{\centering\arraybackslash}p{0.16\textwidth}>{\centering\arraybackslash}p{0.16\textwidth}>{\centering\arraybackslash}p{0.32\textwidth}}
&\multicolumn{5}{c}{\multirow{2}{*}{\includegraphics[width=0.96\textwidth,trim={-0cm 0cm 0.cm 0cm},clip]{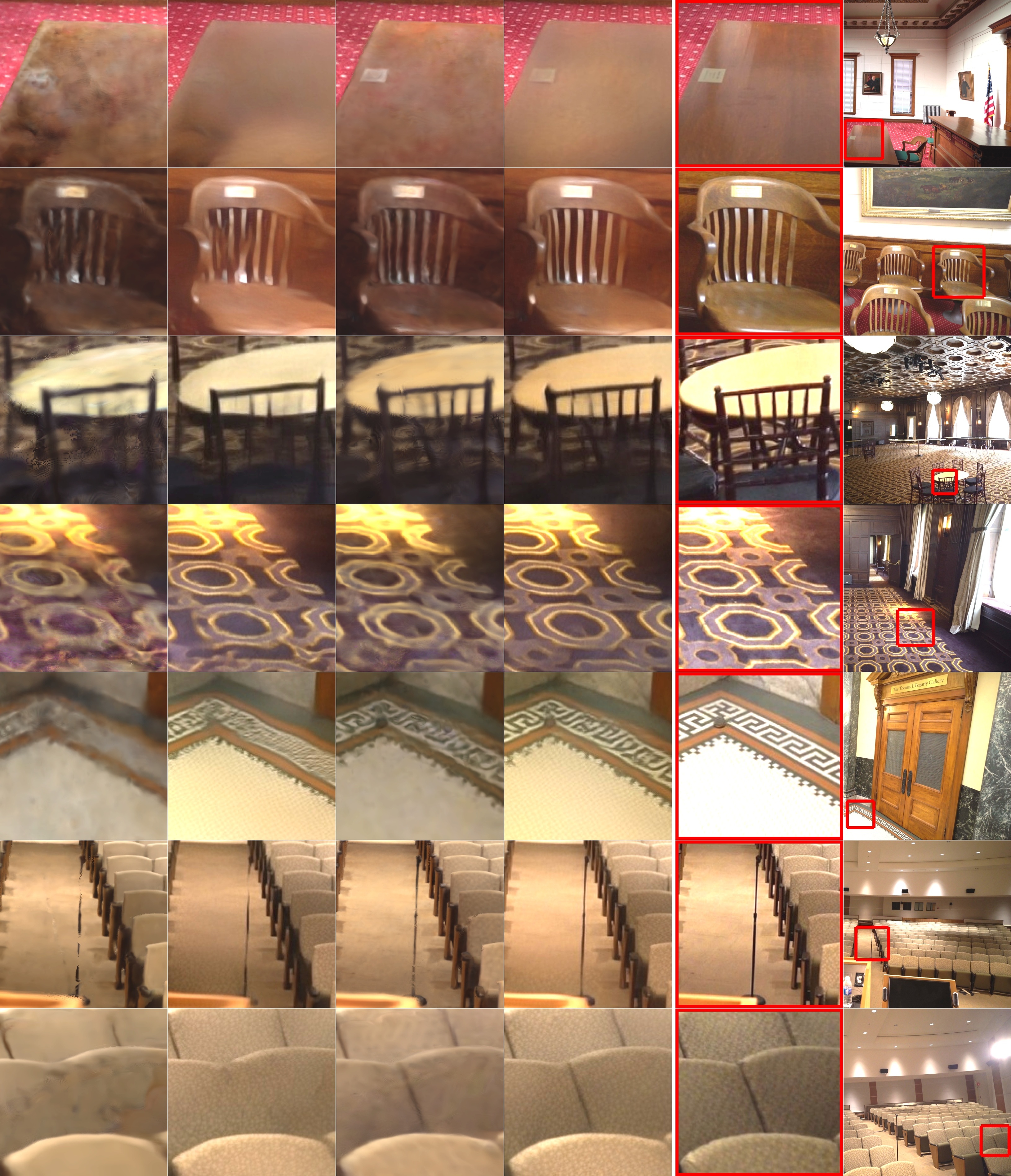}}} \\[24pt]
\large{\textbf{1}} & & & & & \\[71pt]
\large{\textbf{2}}& & & & &\\[71pt]
\large{\textbf{3}}& & & & &\\[71pt]
\large{\textbf{4}}& & & & &\\[71pt]
\large{\textbf{5}}& & & & &\\[71pt]
\large{\textbf{6}}& & & & &\\[71pt]
\large{\textbf{7}}& & & & &\\[35pt]
&Nerfacto & Ours w/o discriminator & Ours w/o generator & Ours  & Ground truth \\
  \end{tabular}
  \vspace{-.2cm}
   \caption{Ablation experiments on Tanks and Temples. ``Ours w/o discriminator'' significantly struggles with patterns (samples 4 and 5) and misses thin structures (samples 3 and 6). ``Ours w/o generator'' better recovers the patterns but produces blurry results compared to the full method (samples 4, 5 and 7).}
   \label{fig:tnt_ablation}
\end{figure*}
\begin{table}[tb]
  \centering
  \caption{Ablation study on Auditorium scene from Tanks and Temples. The decline in performance when removing individual parts of our method confirms our design choices. }
  \label{tab:ablations_tnt}
\resizebox{\linewidth}{!}{
  \begin{tabular}{lcccc}
    \toprule
    Method & PSNR$\uparrow$ & SSIM$\uparrow$ & LPIPS$\downarrow$ & KID $\downarrow$ \\
    \midrule
    Nerfacto \cite{nerfstudio} & 21.1 & 0.843 & 0.304 & 0.0378 \\
    Ours & \textbf{22.3} & \textbf{0.862} & \textbf{0.158} & \textbf{0.0107} \\
    \quad w/o discriminator & 21.9 & 0.857 & 0.178 & 0.0178 \\
    \quad w/o generator & 21.7 & 0.854 & 0.247 & 0.0155 \\
    \quad \textcolor{blue}{NeRF training}  \\
    \quad \quad \textcolor{blue}{w/o adv.\ loss} & 21.9 & 0.861 & 0.175 & 0.0110 \\
    \quad \quad \textcolor{blue}{w/o perc.\ loss} & 21.7 & 0.857 & 0.187 & 0.0194 \\
    \quad \textcolor{purple}{Generator training} \\
    \quad \quad \textcolor{purple}{w/o high-res.\ patches} & 22.1 & 0.851 & 0.179 & 0.0176 \\
    \quad \quad \textcolor{purple}{w/o low-res.\ patches} & 22.1 & \textbf{0.862} & 0.159 & 0.0109 \\
    \quad \quad \textcolor{purple}{w/o RGB encoding} & 22.1 & 0.860 & 0.167 & 0.0122 \\
    \quad \quad \textcolor{purple}{w/o adv.\ loss} & \textbf{22.3} & 0.861 & 0.174 & 0.0122 \\
    \quad \quad \textcolor{purple}{w/o perc.\ loss} & 21.7 & 0.834 & 0.184 & 0.0110 \\
    \quad \quad \textcolor{purple}{w/o RGB loss} & 21.8 & 0.859 & 0.163 & 0.0109 \\
    \bottomrule
  \end{tabular}}
\end{table}
\subsection{Ablation Experiments}
To verify the effectiveness of the added components, we perform an ablation study on the ScanNet++ and Tanks and Temples datasets. The quantitative results (\cref{tab:results_scannet} and \cref{tab:results_tnt}), as well as the qualitative results in \cref{fig:tnt_ablation} show that the complete version of our method achieves the highest performance. 

\subsubsection{Without Discriminator} In this experiment, we investigate the impact of optimizing the radiance field solely with an RGB loss and applying our generator as a pure post-processing step on the renderings. We observe that the geometry and texture details of the scene cannot be recovered to the same extent as achieved with our full method (\cref{fig:tnt_ablation}). The drop in all metrics (``w/o discriminator'' in \cref{tab:results_scannet,tab:results_tnt}) clearly highlights the importance of the patch-based supervision to inform the NeRF representation in 3D. 
This indicates that our method, which incorporates gradient backpropagation to the 3D representation, surpasses a pure 2D post-processing approach. 

\subsubsection{Without Generator} When omitting the generator, the results are less sharp (e.g., samples 4, 5, and 7 in \cref{fig:tnt_ablation}), which leads to lower performance (``w/o generator'' in \cref{tab:results_scannet,tab:results_tnt}) compared to the full version of our method. This shows that the generator helps to achieve high-detail renderings.

\subsubsection{NeRF Training} \cref{tab:ablations_tnt,tab:ablations_scannet} provide more detailed ablations on the NeRF optimization, which show that dropping the adversarial loss (\textcolor{blue}{``w/o adv.\ loss''}) or dropping the perceptual loss (\textcolor{blue}{``w/o perc.\ loss''}) have a negative impact on performance. 

\subsubsection{Generator Training} Detailed ablations on the generator training are listed in \cref{tab:ablations_tnt,tab:ablations_scannet}. 
We investigate the impact of omitting specific input patch resolutions in our model. Removing the two highest resolutions out of the six resolutions results in a noticeable performance drop (\textcolor{purple}{``w/o high-res.\ patches''}), highlighting the essential role of high-resolution input for the generator.
Conversely, when the two lowest resolutions are removed (\textcolor{purple}{``w/o low-res.\ patches''}), we observe a decline in quality  that indicates that the generator also relies on the low-resolution input. We further show that the small RGB encoder, i.e., one convolutional layer on the multi-resolution patches (\cref{fig:sg2}), benefits the generator (\textcolor{purple}{``w/o RGB encoding''}). 

Furthermore, our ablation experiments examine the individual loss functions, namely the adversarial loss (\textcolor{purple}{``w/o adv. loss''}), perceptual loss (\textcolor{purple}{``w/o perc. loss''}), and L1 RGB loss (\textcolor{purple}{``w/o RGB loss''}). The results reveal that each of these loss functions contributes significantly to the overall performance of the generator.

\begin{table}[t]
  \centering
  \caption{Ablation study with reduced number of images on a large ScanNet++ scene. 
  Our method consistently outperforms Nerfacto by a similar margin, regardless of the level of input sparsity.}
  \label{tab:results_scannet_few_images}
\resizebox{\linewidth}{!}{
  \begin{tabular}{@{}lccccc@{}}
    \toprule
    Method & \# images & PSNR$\uparrow$ & SSIM$\uparrow$ & LPIPS$\downarrow$ & KID $\downarrow$ \\
    \midrule
    Nerfacto \cite{nerfstudio}& \multirow{2}{*}{800} & 24.2 & 0.844 & 0.247 & 0.0661 \\
    Ours & & \textbf{24.7} & \textbf{0.870} & \textbf{0.169} & \textbf{0.0157} \\
    \cmidrule(lr){2-2}
    Nerfacto \cite{nerfstudio} & \multirow{2}{*}{400} & 24.2 & 0.843 & 0.252 & 0.0766 \\
    Ours & & \textbf{24.3} & \textbf{0.864} & \textbf{0.169} & \textbf{0.0176} \\
    \cmidrule(lr){2-2}
    Nerfacto \cite{nerfstudio} & \multirow{2}{*}{200} & 23.5 & 0.825 & 0.274 & 0.0867 \\
    Ours & & \textbf{23.9} & \textbf{0.862} & \textbf{0.182} & \textbf{0.0206} \\
    \cmidrule(lr){2-2}
    Nerfacto \cite{nerfstudio} & \multirow{2}{*}{100} & 22.2 & 0.805 & 0.307 & 0.1069 \\
    Ours & & \textbf{22.5} & \textbf{0.842} & \textbf{0.216} & \textbf{0.0289} \\
    \cmidrule(lr){2-2}
    Nerfacto \cite{nerfstudio} & \multirow{2}{*}{50} & 20.2 & 0.771 & 0.347 & 0.1564 \\
    Ours & & \textbf{20.5} & \textbf{0.788} & \textbf{0.282} & \textbf{0.0858} \\
    \cmidrule(lr){2-2}
    Nerfacto \cite{nerfstudio} & \multirow{2}{*}{25} & 15.5 & 0.686 & 0.524 & 0.2432 \\
    Ours & & \textbf{16.7} & \textbf{0.727} & \textbf{0.408} & \textbf{0.1738} \\
    \bottomrule
  \end{tabular}
  }
\end{table}
\begin{table}[tb]
  \centering
  \caption{Ablation study on a ScanNet++ scene. Removing individual parts of our methods leads to a decline in performance, indicating the importance of the different architectural choices and training strategies.}
  \label{tab:ablations_scannet}
\resizebox{\linewidth}{!}{
\begin{tabular}{lcccc}
    \toprule
    Method & PSNR$\uparrow$ & SSIM$\uparrow$ & LPIPS$\downarrow$ & KID $\downarrow$ \\
    \midrule
    Nerfacto \cite{nerfstudio} & 24.2 & 0.844 & 0.247 & 0.0661 \\
    Ours & \textbf{24.7} & \textbf{0.870} & \textbf{0.169} & \textbf{0.0157} \\
    \quad w/o discriminator & 24.4 & 0.859 & 0.188 & 0.0194 \\
    \quad w/o generator & 24.6 & 0.865 & 0.201 & 0.0294 \\
    \quad \textcolor{blue}{NeRF training}  \\
    \quad \quad \textcolor{blue}{w/o adv.\ loss} & 24.4 & 0.867 & 0.170 & 0.0172 \\
    \quad \quad \textcolor{blue}{w/o perc.\ loss} & 24.5 & 0.866 & 0.184 & 0.0229 \\
    \quad \textcolor{purple}{Generator training} \\
    \quad \quad \textcolor{purple}{w/o high-res.\ patches} & 24.2 & 0.849 & 0.202 & 0.0188 \\
    \quad \quad \textcolor{purple}{w/o low-res.\ patches} & 24.6 & \textbf{0.870} & 0.171 & 0.0180 \\
    \quad \quad \textcolor{purple}{w/o RGB encoding} & 24.6 & \textbf{0.870} & 0.171 & 0.0179 \\
    \quad \quad \textcolor{purple}{w/o adv.\ loss} & \textbf{24.7} & 0.869 & 0.175 & 0.0184 \\
    \quad \quad \textcolor{purple}{w/o perc.\ loss} & 24.4 & 0.858 & 0.170 & 0.0217 \\
    \quad \quad \textcolor{purple}{w/o RGB loss} & 24.6 & 0.869 & 0.170 & 0.0172 \\
    \bottomrule
  \end{tabular}}
\end{table}

\subsubsection{Reduced Number of Images}
\cref{tab:results_scannet_few_images} lists results when reducing the number of images of the largest scene (800 images) to 400, 200, 100, 50 and 25 images, which is extremely sparse for room-scale scenes. It shows that our method consistently outperforms Nerfacto by a similar margin as observed in the more dense setting. 

\subsection{Evaluations}
\subsubsection{View Consistency}
\label{sssec:consistency}
\begin{figure}[tb]
 \setlength\tabcolsep{1.25pt}
  \centering
  \begin{tabular}{>{\centering\arraybackslash}p{0.03\linewidth}>{\centering\arraybackslash}p{0.26\linewidth}>{\centering\arraybackslash}p{0.02\linewidth}>{\centering\arraybackslash}p{0.26\linewidth}>{\centering\arraybackslash}p{0.02\linewidth}>{\centering\arraybackslash}p{0.35\linewidth}}
& & & & & \small Pixel column over time \\
& \small First Frame & & \small Last Frame & & \small from \textcolor{cyan}{first} to \textcolor{magenta}{last} frame \textbf{\textcolor{cyan}{|}}$\rightarrow$\textbf{\textcolor{magenta}{|}}\\
&\multicolumn{5}{c}{\multirow{2}{*}{\includegraphics[width=0.97\linewidth,trim={-0cm 0cm 0.cm 0cm},clip]{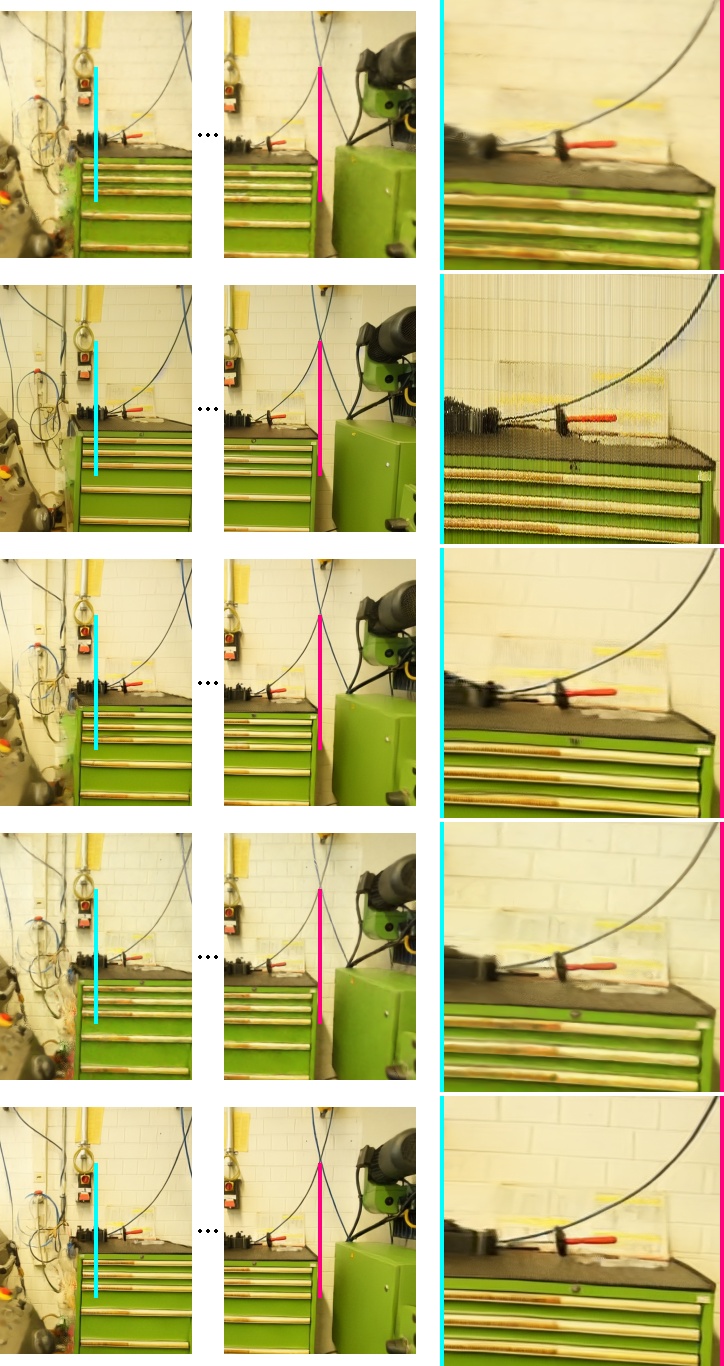}}} \\[-10pt]
\rotatebox[origin=c]{90}{\parbox{3cm}{\centering \small Nerfacto}}&&&&&\\[42pt]
\rotatebox[origin=c]{90}{\parbox{3cm}{\centering \small Nerfacto + ControlNet}}&&&&&\\[40pt]
\rotatebox[origin=c]{90}{\parbox{3cm}{\centering \small Ours w/o discriminator}}&&&&&\\[41pt]
\rotatebox[origin=c]{90}{\parbox{3cm}{\centering \small Ours w/o generator}}&&&&&\\[41pt]
\rotatebox[origin=c]{90}{\parbox{3cm}{\centering \small Ours}}&&&&&\\
  \end{tabular}
   \caption{View consistency visualization. From a sequence of video frames, we extract a column of pixels at the same position in each frame (left) and concatenate them horizontally to visualize view consistency (right). }
   \label{fig:video_consistency}
\end{figure}

To obtain view-consistent results, it is essential to first backpropagate to the NeRF and optimize the underlying 3D representation. Following~\cite{Lai-ECCV-2018}, \cref{tab:view_consistency} shows consistency of test views computed with optical flow~\cite{raft}. Comparing Nerfacto and our method without backpropagation to NeRF (i.e., ``w/o discriminator''), shows that our generator largely improves visual quality (i.e., KID), while adding marginal inconsistencies. However, our full method backpropagates to the 3D representation, thus reducing the added inconsistencies by over half. Our video results show that inconsistencies are barely noticeable, and qualitative improvements dominate. The alternative refinement with ControlNet (Nerfacto + ControlNet) achieves high visual quality, however, the refined views are highly inconsistent. Following \cite{wang20224k}, \cref{fig:video_consistency} visualizes view consistency by tracking a column of pixels across a sequence of video frames: the inconsistency of Nerfacto + ControlNet is clearly visible as vertical stripe patterns. The full version of our method shows the best combination of view consistency and visual quality in \cref{fig:video_consistency}.

\begin{table}[tb]
  \centering
\caption{View consistency evaluation on ScanNet++. Novel views from our method are both highly view-consistent and of high visual quality (i.e., KID). \textbf{Best} and \underline{second best} results are highlighted. }
\label{tab:view_consistency}
\resizebox{\linewidth}{!}{
  \begin{tabular}{l@{}cc}
    \toprule
    Method & View Consistency MSE $\downarrow$ & KID $\downarrow$ \\
    \midrule
    Nerfacto \cite{nerfstudio} & \textbf{0.0018} & 0.0398 \\
    Nerfacto + ControlNet \begin{minipage}{0.21\linewidth}\small \raggedright \linespread{0.}\selectfont \cite{zhang2023adding}\end{minipage} & 0.0039 & \textbf{0.0097} \\
    Ours w/o discriminator & 0.0023 & 0.0143 \\
    Ours w/o generator & \textbf{0.0018} & 0.0169 \\
    Ours & \underline{0.0020} & \underline{0.0113} \\
    \bottomrule
  \end{tabular}}
\end{table}

\subsubsection{Alternative NeRF Backbone}
\label{sssec:nerf_backbone}
\begin{table}[tb]
  \centering
  \caption{Using different NeRF backbones on ScanNet++. Our method is flexible to use an alternative NeRF representation, e.g., Instant NGP. The default approach builds on Nerfacto.}
  \label{tab:nerf_backbone}
\resizebox{\linewidth}{!}{
  \begin{tabular}{lcccc}
    \toprule
    Method & PSNR$\uparrow$ & SSIM$\uparrow$ & LPIPS$\downarrow$ & KID $\downarrow$ \\
    \midrule
    Instant NGP \cite{mueller2022instant} & 25.3 & 0.844 & 0.269 & 0.0511 \\
    Ours w/ Instant NGP & \textbf{25.6} & \textbf{0.857} & \textbf{0.173} & \textbf{0.0148} \\
    \cmidrule(lr){1-5}
    Nerfacto \cite{nerfstudio} & 25.6 & 0.848 & 0.245 & 0.0398 \\
    Ours & \textbf{26.1} & \textbf{0.864} & \textbf{0.161} & \textbf{0.0113} \\
    \bottomrule
  \end{tabular}
  }
\end{table}
\begin{table}[tb]
  \centering
\caption{Runtime comparison on ScanNet++. Our rendering time is competitive with Nerfacto; the training time is likely comparable to Mip-NeRF 360 which was trained on 4 GPUs vs.\ our method was trained on a single GPU.}
\label{tab:runtime}
\resizebox{\linewidth}{!}{
  \begin{tabular}{@{}l@{}r@{\,\,}c@{\;\;}r@{}c@{}}
    \toprule
    Method & \multicolumn{2}{c}{Training Time $\downarrow$} & \multicolumn{2}{c}{Rendering Time $\downarrow$} \\
    & & & \multicolumn{2}{c}{per frame (768x1152)} \\
    \midrule
    Mip-NeRF 360 \cite{barron2022mipnerf360} & 18\,h & \small 4x NVIDIA V100 & 24\,s & \small 1x NVIDIA V100 \\
    Instant NGP \cite{mueller2022instant} & 3.8\,h & \multirow{3}{*}{\rotatebox[origin=c]{90}{\parbox{1.cm}{\centering \small 1x NVIDIA RTX A6000}}} & \textbf{0.47}\,s & \multirow{3}{*}{\rotatebox[origin=c]{90}{\parbox{1.cm}{\centering \small 1x NVIDIA RTX A6000}}} \\
    Nerfacto \cite{nerfstudio} & \textbf{3.5}\,h & & 0.87\,s & \\
    Ours & 2\,d 10\,h & & 0.89\,s & \\
    \bottomrule
  \end{tabular}}
\end{table}
We build our method on the Nerfacto radiance field representation, however, it is flexible to backpropagate gradients to a different NeRF backbone. Using Instant NGP as alternative NeRF, leads to improvements of similar magnitude across all metrics compared to using Nerfacto (\cref{tab:nerf_backbone}). 

With the Instant NGP backbone, we use the parameters as described in \cref{ssec:train_details}, except, due to higher memory consumption of the Instant NGP rendering, we reduce the patch size to $192\times 192$, from which we crop four $64 \times 64$ patches for the discriminator. The loss weights are then set to $\lambda^\mathtt N_{\mathrm{adv}}=0.001$, $\lambda^\mathtt N_{\mathrm{perc}}=0.001$ and $\lambda^\mathtt N_{\mathrm{gp}}=0.1$. 
\subsubsection{Runtime}
\cref{tab:runtime} provides a comparison of training and rendering times. The rendering time of our method is dominated by the rendering time of the underlying NeRF, since the generator forward pass is comparably fast, i.e., takes only ~13ms. Hence, our method renders novel views at effectively the same speed as Nerfacto and much faster than Mip-NeRF 360. Using Instant NGP as alternative NeRF backbone (\cref{sssec:nerf_backbone}) speeds up rendering towards the Instant NGP rendering time. 
Our training time is slower than Nerfacto or Instant NGP; however, our method significantly improves visual quality. All methods are trained until convergence. 
\subsubsection{Impact of View Coverage on Visual Quality}
\begin{figure}[tb]
\centering
\includegraphics[width=\linewidth,trim={0.2cm 0.2cm 0.2cm 0.2cm},clip]{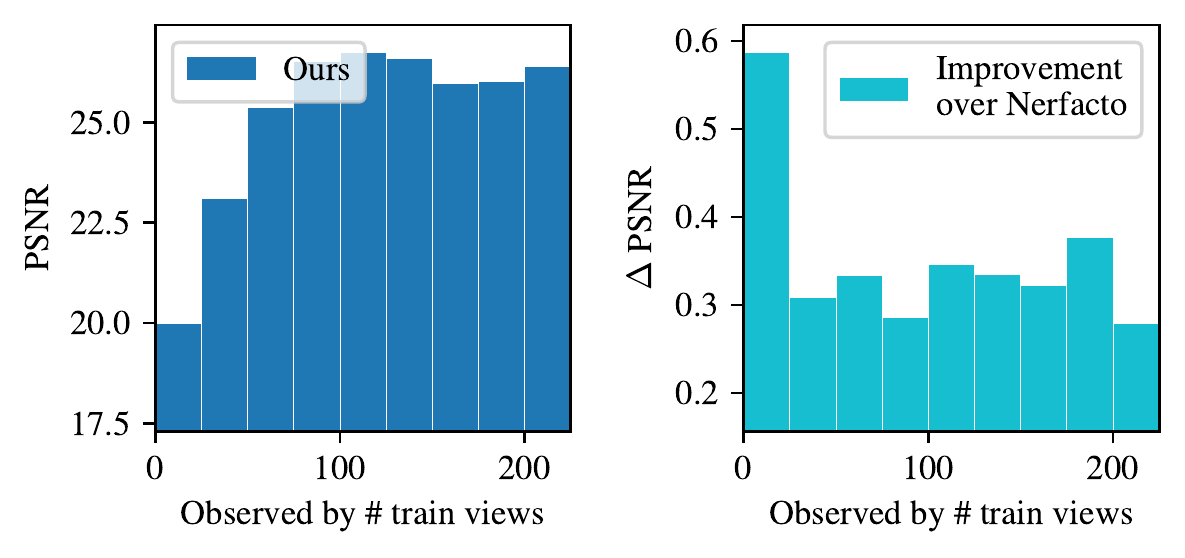}
\caption{Impact of view coverage on PSNR (left)/ improvement over Nerfacto (right) on a ScanNet++ scene. View coverage is computed using ground truth depth and counting \# train views that project to pixels in test views.}
\label{fig:quality_coverage}
\end{figure}
\begin{figure}[tb]
 \setlength\tabcolsep{0pt}
  \centering
  \begin{tabular}{>{\centering\arraybackslash}p{0.33\linewidth}>{\centering\arraybackslash}p{0.33\linewidth}>{\centering\arraybackslash}p{0.15\linewidth}>{\centering\arraybackslash}p{0.15\linewidth}>{\centering\arraybackslash}p{0.03\linewidth}}
& & {\footnotesize $\Delta$ LPIPS:} {\small $\;0$} & \includegraphics[width=\linewidth,trim={0cm 0.3cm -0.5cm 0cm},clip]{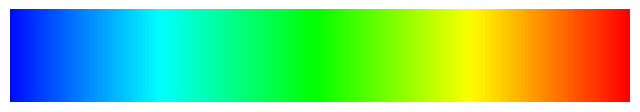} & \small $0.2$ \\
\multicolumn{5}{c}{\includegraphics[width=\linewidth,trim={0cm 0cm 0cm 0cm},clip]{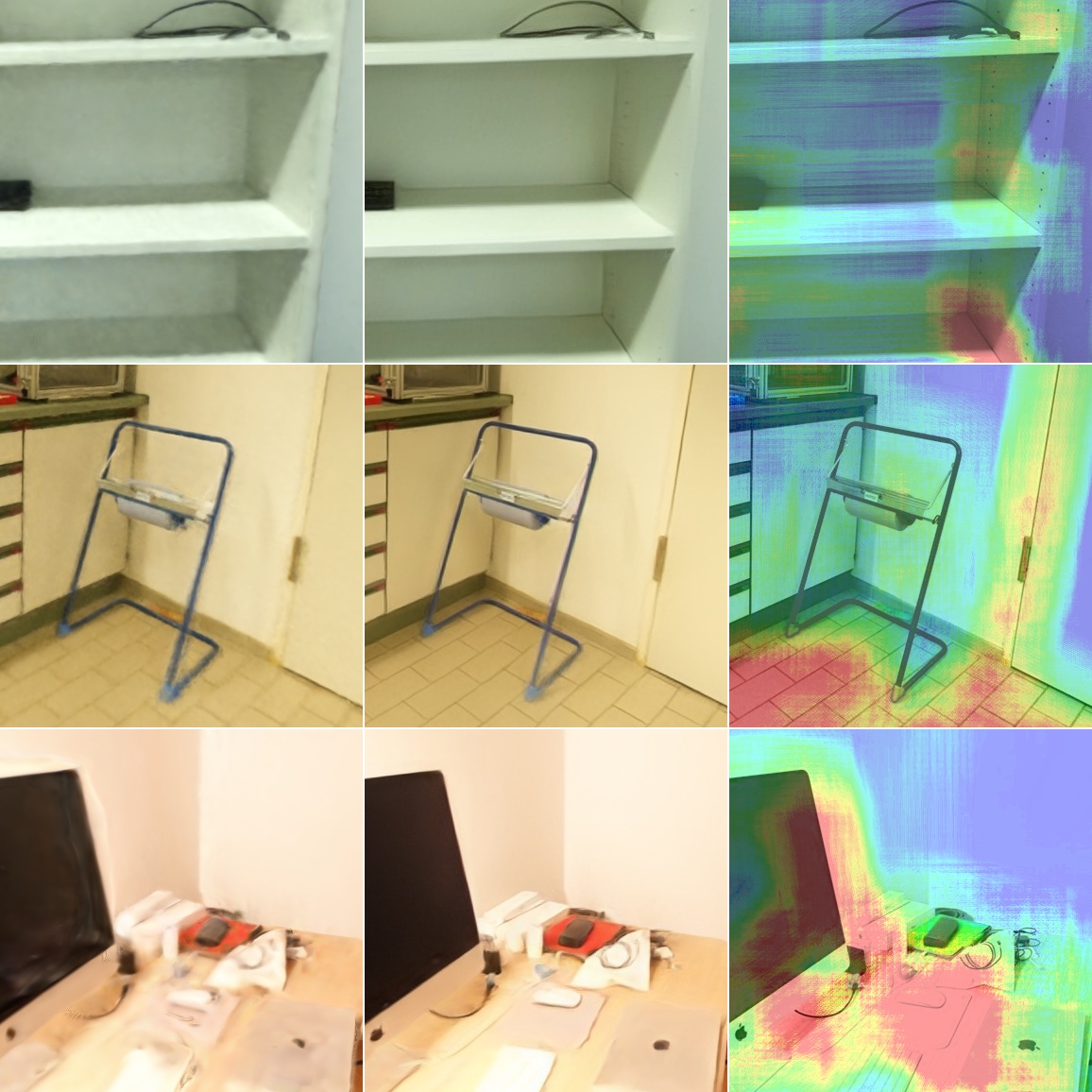}}\\
Nerfacto & Ours & \multicolumn{3}{c}{\small Ground truth overlayed} \\
& & \multicolumn{3}{c}{\small w/ LPIPS improvement} \\
  \end{tabular}
   \caption{Perceptual improvement over Nerfacto. Our method particularly improves structured areas with objects or floor patterns, compared to texture-less surfaces. LPIPS is computed on patches.
   }
   \label{fig:lpips_improvement}
\end{figure}
\cref{fig:quality_coverage}-left shows that PSNR of novel views increases with view coverage, and stagnates for areas that have been observed by $>75$ training views. The improvement of our method over Nerfacto (\cref{fig:quality_coverage}-right) is twice as high in regions of low coverage, that are observed by $<25$ training views, compared to areas of higher coverage. \cref{fig:lpips_improvement} further visualizes the improvement in terms of LPIPS by calculating LPIPS on small $96 \times 96$ patches of test views in a convolutional manner. It shows that improvements particularly happen in richly structured areas, such as table-top objects or repetitive floor patterns, rather than texture-less walls. 
\subsubsection{GAN Hallucination Effects}
\begin{figure}[tb]
 \setlength\tabcolsep{1.25pt}
  \centering
  \begin{tabular}{>{\centering\arraybackslash}p{0.5\linewidth}>{\centering\arraybackslash}p{0.5\linewidth}}
\multicolumn{2}{c}{\includegraphics[width=\linewidth,trim={0cm 0cm 0cm 0cm},clip]{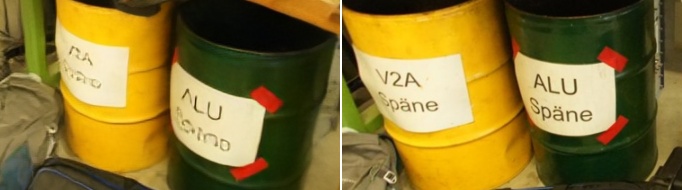}} \\
Ours & Nearby training view \\
  \end{tabular}
   \caption{GAN hallucination effects. Reconstruction of inscriptions is difficult, which can cause hallucinated characters (left).}
   \label{fig:gan_hallucination}
\end{figure}
Using generative approaches for reconstruction comes with the danger of hallucinating content, which may not be view-consistent. 
Through backpropagation to the NeRF representation and the generator conditioning, our novel views are highly view-consistent (\cref{sssec:consistency}) and closely match the ground truth, e.g., difficult patterns in \cref{fig:tnt_baseline_comp_gen}. 
We, however, observe that reconstructing written text is very challenging, which can lead to hallucinated, incorrect characters as shown in \cref{fig:gan_hallucination}.
\subsubsection{Discriminator on Unseen Views} Rendering fake patches from sampled poses rather than the training views, has potential to provide additional discriminator feedback to the NeRF representation. Our attempt to sample poses from Gaussians around the training poses to render fake patches, however, left the average performance unchanged. Likely, a more sophisticated strategy is needed to sample patches that benefit from additional supervision. 
\subsection{Limitations}
Our results show that we can achieve significant improvements compared to state-of-the-art methods.
At the same time, we believe there are still important limitations.
For instance, at the moment our patch discriminator is trained per scene. 
However, it would be beneficial to train a more generic prior that has access to a larger scene corpus to improve its capability.
While feasible, a naive generalizable discriminator would tend to collapse to a scene classifier; i.e., it would primarily identify whether a patch belongs to the same scene or not. 
To overcome this issue, one possible solution is to train a NeRF representation for each training scene simultaneously, which would entail considerable computational expenses.
Another limitation is that we currently focus only on static scenes; however, we believe it would be interesting to expand our approach to recent deformable and dynamic NeRF approaches such as \cite{park2021hypernerf, tretschk2021nonrigid, kirschstein2023nersemble, isik2023humanrf}. 

\section{Conclusion}
We introduce \OURS, a new approach for adversarial optimization of neural radiance fields.
The main idea behind our approach is to impose patch-based rendering constraints into the radiance field reconstruction.
By backpropagating gradients from a scene patch discriminator, we effectively address typical imperfections and rendering artifacts stemming from traditional NeRF methods.
In particular, in regions with limited coverage, this significantly improves rendering quality, both qualitatively and quantitatively outperforming state-of-the-art methods.
At the same time, our method is only a stepping stone for combining rendering priors with high-end novel view synthesis. 
For instance, we believe that generalizing across scenes will offer numerous opportunities for leveraging similar ideas as those proposed in this work.
\begin{acks}
This work was funded by a Meta SRA. Matthias Nießner was also supported by the ERC Starting Grant Scan2CAD (804724).
We thank Angela Dai for the video voice-over. 
\end{acks}
\bibliographystyle{ACM-Reference-Format}
\bibliography{bibliography}
\end{document}